\documentclass[letterpaper, 10 pt, conference]{ieeeconf}

\usepackage{times}  
\usepackage{helvet} 
\usepackage{courier}  
\usepackage[hyphens]{url}  
\usepackage{graphicx} 
\usepackage{xcolor}

\usepackage{soul}
\usepackage[utf8]{inputenc}
\usepackage{booktabs}
\usepackage{algorithmic}
\usepackage{mathtools}
\usepackage{amsmath} 
\usepackage{amssymb}  
\usepackage{multirow}

\usepackage[ruled,noend,linesnumbered]{algorithm2e}
\usepackage{standalone}
\usepackage{amsfonts}
\usepackage[mathscr]{euscript}
\let\euscr\mathscr \let\mathscr\relax

\usepackage{accents}

\usepackage{stackengine}

\SetAlFnt{\footnotesize}
\SetAlCapFnt{\footnotesize}
\SetAlCapNameFnt{\footnotesize}

\SetCommentSty{mycommfont}

\newcommand\munderbar[1]{%
  \underaccent{\bar}{#1}}

\title{\LARGE \bf Theory and Analysis of Optimal Planning over Long and Infinite Horizons for Achieving Independent Partially-Observable Tasks that Evolve over Time}

\author{Anahita Mohseni-Kabir$^{1}$, Manuela Veloso$^{1}$, and Maxim Likhachev$^{1}$

\thanks{*This work was partially supported by Sony AI.}
\thanks{$^{1}$The authors are with School of Computer Science, Carnegie Mellon University {\tt\small \{anahitam, mmv, maxim\}@cs.cmu.edu}.}
}

\IEEEoverridecommandlockouts

\begin{document}

\maketitle
\thispagestyle{empty}
\pagestyle{empty}

\begin{abstract}
We present the theoretical analysis and proofs of a recently developed algorithm that allows for optimal planning over long and infinite horizons  for achieving multiple independent tasks that are partially observable and evolve over time.


\end{abstract}

\section{Background}
\vspace{-0.05cm}
\noindent We focus on the class of problems with $N$ independent tasks that evolve over time, proposed in~\cite{mohseni2020efficient}. The robot should interleave its actions to find an optimal sequence of actions to attend to all the tasks. A POMDP that includes a single task and a robot is called a \textit{client POMDP}. The $N$ \textit{client POMDPs} are combined into one large POMDP model called an \textit{agent POMDP}. 
To compute an optimal solution for all the tasks, one should solve the agent POMDP optimally. We briefly describe the client  and agent POMDPs, and how~\cite{mohseni2020efficient} referred to as \textit{multi-task planner} solves the agent POMDP efficiently.


\vspace{-0.15cm}
\subsection{Client POMDP}
\noindent The client POMDP for task $i$ is represented as a tuple $(S_i, A_i, Z_i, T_i, O_i, R_i, \gamma, H)$. The state space $S_i = SR \times SC_i$ includes the robot's state, $SR$, and the other state variables that are specific to task $i$, $SC_i$. The action space $A_i$ includes the actions that can be applied by the robot to task $i$ and a special \textit{no op} action which bears no direct effect on the task (\emph{i.e}., when executed, task $i$ follows its underlying Hidden Markov Model). $Z_i$ denotes the observation space. The robot takes an action $a \in A_i$ and transitions from a state $s \in S_i$ to $s' \in S_i$ with probability $T_i (s, a, s')$. The robot then observes $z \in Z_i$ and receives a reward $R_i(s_i,a)$. 
The function $O_i(s',a, z)$ models the robot's noisy  observations. 

In POMDP planning, the robot keeps a distribution over the states, called a belief state, and searches for a policy $\pi: B_i \rightarrow A$ that maximizes $\mathbb{E}\left[\sum _{{t=0}}^{H}\gamma ^{t}r_{i,t}\right]$ at each belief $b \in B_i$,  where $r_{i,t}$ is the reward gained at time $t$ from POMDP $i$, $H$ is the planning horizon, and $\gamma$ is the discount factor. For infinite-horizon problems with discounting, $H=\infty$ and $0 \leq \gamma < 1$. For finite-horizon problems, $H$ is finite and $\gamma=1.0$. Optimal value $V^*_{i}(b)$ can be computed by iteratively applying the Bellman equation. Similarly, the value of following a  trajectory $n$, $V^{n}_{i}(b_i)$, can be computed by iteratively applying the Bellman equation and following the remaining trajectory~\cite{cassandra1998survey}. In our algorithms, $n$ refers to a trajectory consisting of \textit{no op} actions so $V^{n}_{i}(b_i) \leq V^*_{i}(b_i)$.


\vspace{-0.15cm}
\subsection{Agent POMDP}
\label{sec:agentpomdp}
\noindent A POMDP created from $N$ client POMDPs is called \textit{agent POMDP} (or \textit{robot POMDP}). Let $P = \{i \in \mathbb{N}: i \leq N\}$. Formally, the agent POMDP is represented by $(N, S, A, Z, T, O, R, \gamma, H)$ where $S = SR \times SC_1 \times \ldots \times SC_N$, $A$ and $Z = Z_1 \times \ldots \times Z_N$  denote the state, action and observation spaces respectively. $T$, $O$ and $R$ are the transition, observation and reward functions respectively. The robot's action set $A$ (Eq.~\ref{eq:A_valid}) contains vectors of length $N$ in which except one element, all other elements are \textit{no ops}. The robot's distribution over the states is $b \in B$ where $B = B_1 \times \ldots \times B_N$. The agent POMDP's reward function is additive in terms of its underlying tasks $\mathbb{E}\left[\sum_{{i=1}}^{N} R_{i}\right]$. 

\vspace{-0.45cm}
\small  \begin{equation} \label{eq:A_valid}
\begin{split}
A=\underset{i \in P}\cup\underset{{a \in A_i}}\cup \overbrace{[\text{\small{no op}}...\text{\small{no op}},\underbrace{a}_{i\text{th element}},\text{\small{no op}}...\text{\small{no op}}]}^{\text{length N}}
\end{split}
\end{equation}  \normalsize
\vspace{-0.3cm}

\noindent Using the mathematical definition of the independent tasks, the optimal value of an agent POMDP built from a set of independent client POMDPs $P$ can be iteratively computed as follows where $\Pr(z|b,a) = \prod_{k \in P}{\Pr(z_k|b_k,a[k])}$~\cite{mohseni2020efficient}.


\vspace{-0.3cm}
\small  \begin{equation} \label{eq:agent_POMDP_value}
\begin{split}
V^{*}_{t}(b)&=\max_{a\in A}{\Bigl [}\sum\limits_{{i\in P}}\sum\limits_{{s\in S_i}}b_i(s)R_i(s,a[i]) \\
&+ \gamma \sum\limits_{z\in Z} \Pr(z|b,a) V^*_{t-1}(b_z^a){\Bigr ]} \\
\end{split} \raisetag{2\baselineskip}
\end{equation}  \normalsize
\vspace{-0.35cm}

\noindent To compute the value of the current belief state of the robot for a fixed horizon $H$, the planner starts with the robot's current belief as the root of a tree and builds the tree of all reachable beliefs by considering all possible actions and observations. For each action and observation, a new belief node is added to the tree as a child node of its immediate previous belief. To solve the agent POMDP, a combined belief tree of all tasks is built till horizon $H$. The value of the robot's current belief is computed by propagating value estimates up from the fringe nodes, to their ancestors, all the way to the root, according to Eq.~\ref{eq:agent_POMDP_value}. We call this approach \textit{agent POMDP with a fixed horizon} or \textit{agent-POMDP-FH}.


\vspace{-0.13cm}
\subsection{Multi-task POMDP Planner (Multi-task-FH)}
\label{sec:multitaskFH}

\noindent Note that the agent POMDP approach is impractical if the number of tasks are large. If each client POMDP has $|S|$ states and $|A|$ actions (excluding the \textit{no op} action), and there are $N$ tasks, the robot should plan over an agent POMDP with $|S|^N$ states and $N\times|A|+1$ actions which is infeasible.
Our prior work exploits the observation that in some domains the number of tasks, $k^*$, that the robot can attend to within $H$ is limited~\cite{mohseni2020efficient}. 
Given this observation, if the robot optimally solves all possible sub-problems of size $k^*$ with different combinations of tasks, it can find the optimal solution to the agent POMDP. In~\cite{mohseni2020efficient}, we prove that decomposing  the  agent POMDP into a series of sub-problems of size $k^*$ and solving all combinations of $k^*$ out of $N$ tasks, $tpls = \{tpl \in \euscr{P} (P):|tpl| = k^*\}$ and returning the action with the highest value from them is the optimal solution to the agent POMDP. Symbol $\euscr{P}$ represents the power set.
Note that each member $tpl$ (tuple of size $k^*$) of the set $tpls$ is a sub-problem that can be solved by building a combined POMDP from the POMDPs in $tpl$. The robot assumes that a trajectory of \textit{no op} actions is being executed on the POMDPs that are not in $tpl$. $k^*$ is provided to the algorithm, but we discuss a way to compute it in~\cite{mohseni2020efficient}. 

The prior work uses the solutions to the individual client POMDPs to compute lower and upper-bounds on the optimal value of the agent POMDP  to prune the $tpls$ set. Different from~\cite{mohseni2020efficient} that only uses the solutions to the single tasks to prune the low-quality tasks, in this work we take a more gradual approach and monotonically improve the bounds to prune the low-quality tasks. We start with single tasks ($k=1$), but gradually increase $k$ and solve sub-problems of size $k$ ($<k^*$) to prune the tasks. Since our algorithm gradually improves the bounds to prune as many tasks as possible, it eventually solves less number of sub-problems of size $k^*$ compared to~\cite{mohseni2020efficient}. We first use the single tasks to prune, then pairs, then triplets, and so on. In addition, we use a truncated horizon $h$ ($h < H$) to compute the solutions to the sub-problems of size $k$ rather than the full horizon $H$ which is needed to solve the sub-problems of size $k^*$. This gradual and monotonic improvement of the bounds and planning until a truncated horizon $h$ enables the robot to efficiently and optimally plan over \emph{long} fixed-length horizons without discounting and \emph{infinite-length} horizons with discounting rather than planning for a short fixed horizon as done in~\cite{mohseni2020efficient}.

\vspace{-0.1cm}
\section{Approach}
\vspace{-0.05cm}

\noindent In this section, we first explain the main ideas that we use to extend the agent POMDP planner (explained in section~\ref{sec:agentpomdp}) to be applied on long-horizon problems. We call this new approach \textit{agent POMDP with adaptive horizon} or \textit{agent-POMDP-AH}. We then explain how the agent-POMDP-AH is extended to include the key insights and the efficiency of~\cite{mohseni2020efficient} (multi-task-FH explained in section~\ref{sec:multitaskFH}). We call our approach \textit{multi-task POMDP with adaptive horizon} (\textit{multi-task-AH}) since in addition to leveraging the multiple independent tasks structure, we adapt the horizon (specifically, iteratively increase it) to improve the solution's quality. 




Similar to~\cite{mohseni2020efficient}, we use an online planning framework which interleaves planning and execution. Its main loop is in Alg.~\ref{alg:main}.  During the planning phase, the algorithm computes the best action to execute given the robot's current belief (lines 3-7). In the execution phase, the robot executes the selected action (line 8), updates the belief state (line 9), and replans after each action execution.

\subsection{Agent POMDP with Adaptive Horizon}
\vspace{-0.05cm}
\noindent We adapt the agent-POMDP-FH approach for the class of problems with multiple independent tasks to enable the robot to efficiently plan for long horizons. 
This approach uses a similar procedure to solve the agent POMDP as agent-POMDP-FH, but modifies it with two main ideas. The key ideas are that instead of expanding the belief tree of all the tasks for the full horizon $H$, the robot 1) builds the belief tree until a truncated but gradually increasing horizon $h$ and 2) computes the lower and upper-bounds on the value of the fringe nodes at the truncated horizon. 
To compute the bounds for the fringe nodes, the robot only solves the individual tasks for the remaining horizon $H-h$ (or $\infty$ in the infinite-horizon case) and combines their solutions. It then computes the lower and upper-bounds for the non-fringe nodes by propagating the bound values up from the fringe nodes by following the Bellman equation in Eq.~\ref{eq:agent_POMDP_value}. Note that when planning with a truncated horizon $h$, the planner expands the combined model of all the tasks only till the truncated horizon $h$, \ul{but the individual tasks are solved till the full horizon $H$ to compute the bounds}. 
When the lower and upper-bounds on the value of the robot's belief become equal, the optimal solution is found and the search is terminated. This enables the robot to terminate the search before reaching the full planning horizon $H$.
We call the agent POMDP solver that follows this process \texttt{TruncatedAgentPOMDP}. For long horizons, solving the individual tasks (to compute the bounds) is much faster than expanding the belief tree of the combined model; thus, this approach is efficient compared to the agent-POMDP-FH.

Instead of planning for a fixed horizon $H$, this algorithm (Alg.~\ref{alg:main}) performs planning for increasing values of horizon $h$ until one of the following conditions are satisfied: 1) the horizon limit $H$ is reached, or 2) the lower-bound $\munderbar{V}$ on the value of the robot's belief is equal to its upper-bound $\bar{V}$ (line 4). The first condition assures that the algorithm is terminated when it reaches the maximum  horizon $H$ and outputs the same solution as planning for a fixed horizon $H$. 
The second condition enables the robot to terminate planning before reaching the full planning horizon, thus being more efficient than the agent POMDP approach with a fixed horizon  $H$. 






Alg.~\ref{alg:agent} provides the implementation of some of the functions in Alg.~\ref{alg:main} for the agent-POMDP-AH approach. The \texttt{TruncatedAgentPOMDP} solver builds a combined model with all the client POMDPs in $P$. It finds the bounds for the fringe nodes using the \texttt{ComputeBounds} function and propagates the bounds up to compute the bounds for the non-fringe nodes. We refer to all the POMDPs in $tpl$ as ${tpl}_u$; for the agent POMDP, ${tpl}_u=P$ (all possible tasks). The intuition behind the lower-bound computation (line $7$) is to only consider the best client POMDP from $tpl$ and perform \textit{no ops} on the other POMDPs. This is similar to taking a greedy approach of always selecting the best task to attend to rather than interleaving the tasks. This is indeed a possible solution, hence it is the lower-bound. 
The intuition behind the upper-bound computation is to assume that the robot can address all the client POMDPs (tasks) in $tpl$ in parallel. We only have one robot so this is an upper-bound.



Since the client POMDPs are solved over and over for different beliefs and horizons during planning to compute the bounds, their solutions are cached and reused in the process.






\vspace{-0.4cm}
\begin{algorithm}
	\caption{Online Planner with Adaptive Horizon}
	\label{alg:main}
	\SetKwFunction{FMain}{}
	\SetKwProg{Fn}{MultiTaskAdaptiveHorizonPlanner}{}{}
	\Fn{\FMain{env, P, h, H}}{

		\While{\textbf{not} {AllTasksDone()}}{
		  tpls $\gets$ InitializeTuples(P,h,H) \\
		    \While{{ $\munderbar{V} \neq \bar{V}$ or $h \neq H$}}{			    {a},tpls,{$\munderbar{V}$},{$\bar{V}$} $\gets$ SelectAction({P},{h},{H},tpls) \\
		    h $\gets$ h+1 \\
		    tpls $\gets$ RecomputeTuples(h,tpls) \tcp*[h]{this function is only needed in the multi-task-AH approach}
			}
			{observations} $\gets$ Step({env}, {a})
	
		 UpdateBeliefs(P,{observations})\label{alg:aed-updatepolicy}
		}
	}
\end{algorithm}
\vspace{-0.7cm}
\vspace{-0.3cm}
\begin{algorithm}
	\caption{Agent POMDP with Adaptive Horizon}
	\label{alg:agent}
	\SetKwFunction{FMain}{}
    \SetKwProg{Fn}{InitializeTuples}{}{}
    \Fn{\FMain{$P$,$h$,$H$} \Return $tpls$ $\gets$ \{($P$,$\emptyset$)\} }{
    	   
    }
	\SetKwFunction{FMain}{}
    \SetKwProg{Fn}{SelectAction}{}{}
    \Fn{\FMain{$P$,$h$,$H$,$tpls$}}{
    	($\munderbar{V}_{P}$,$\bar{V}_{P}$) $\gets$ TruncatedAgentPOMDP($h$,$H$,$tpls$) \\
    	$a_{best} \gets$ action with highest $\bar{V}_{P}$ \\
    	\Return $a_{best}$,$tpls$,$\munderbar{V}_{P}$,$\bar{V}_{P}$
    	}
    \SetKwFunction{FMain}{}
    \SetKwProg{Fn}{ComputeBounds}{\tcp*[h]{for the remaining horizon $H-h$}}{}
    \Fn{\FMain{$b$,$tpl$}}{ 
    	    $ \munderbar{V} \gets \smashoperator{\max_{p\in {{tpl}_u}}} (V_p^*(b_p) + \smashoperator{\sum\limits_{{q\in {{tpl}_u} \setminus \{p\}}}} V_q^{n}(b_q))$; 	$\bar{V} \gets \smashoperator{\sum\limits_{{p\in {{tpl}_u}}}} V_p^{*}(b_p)$ \\
    	    \vspace{-0.1cm}
    	    \Return $\munderbar{V}$,$\bar{V}$
    }
\end{algorithm}
\vspace{-0.64cm}

\subsection{Multi-task POMDP with Adaptive Horizon}
\vspace{-0.02cm}
\noindent We exploit the two key ideas from the previous section and extend the multi-task-FH~\cite{mohseni2020efficient} to address long horizon planning. The multi-task-FH is able to leverage the independent tasks structure in the problem to efficiently solve the agent POMDP, and  agent-POMDP-AH  speeds up planning for long horizons by terminating the search earlier through the truncated horizon and bound computations. We combine the benefits of the two approaches in the multi-task-AH.

Multi-task-FH exploits the observation that within a fixed horizon $H$, the robot can only consider a limited number of tasks $k^*$. Similarly here, we also consider all possible subsets of size $k^*$ as it is needed to ensure optimality. However, in addition to this, we leverage the observation that within the truncated horizon $h$, $h \leq H$, the robot can only consider $k$ tasks ($k \leq k^*$), and it performs \textit{no ops} on the other tasks. Intuitively, we use the key idea of~\cite{mohseni2020efficient} twice, once to divide the agent POMDP of size $P$ into smaller problems of size $k^*$, and the second time to divide the smaller problems of size $k^*$ into sub-problems of size $k$, $k \leq k^*$, that can be solved more efficiently. Leveraging the truncated horizon to further limit the number of tasks that the robot can attend to enables us to significantly speed up planning. The robot only considers combined models of size $k$ till horizon $h$, rather than combined models of size $k^*$, but computes the lower and upper-bounds on $k^*$ individual tasks for the remaining horizon $H-h$. Note that the lower and upper-bound computations are done on the individual tasks till the full horizon $H$; so their computations should consider all $k^*$ tasks to ensure similar optimality guarantees as~\cite{mohseni2020efficient} (as the agent can attend to $k^*$ tasks within horizon $H$).
Our approach is especially powerful in the infinite-horizon problems with $N$ tasks. In such problems the number of tasks that the robot can attend to within $H=\infty$ is $N$, $k^*=N$; thus, if $k \ll N$, the algorithm significantly expedites planning by solving multiple sub-problems of smaller sizes rather than solving the agent POMDP with all the $N$ tasks.
As we increase the truncated horizon $h$, we might need to increase the size of the subsets, \emph{i.e.}, increase $k$. We explain how we address this important aspect of the problem later.

Alg.~\ref{alg:acpomdp} shows the multi-task-AH algorithm. The function \texttt{InitializeTuples} considers all possible subsets of $P$ with size $k^*$ (line 2) and further divides it into subsets of size $k$ (line 3). Each subset of size $k^*$ ($tpl \in tpls$) is divided into two sets, $tpl_c$ with size $k$ and $tpl_l$ with size $k^*-k$. 
The truncated agent POMDP is built from the POMDPs in $tpl_c$ while executing \textit{no ops} on the POMDPs in $tpl_l$, but the bound computations for the fringe nodes are done on all POMDPs in $tpl_u=tpl_c \cup tpl_l$ to assure valid lower and upper-bounds on the value of the tuple $tpl$ (\texttt{TruncatedAgentPOMDP} function). 
The \texttt{SelectAction} function solves a truncated agent POMDP for each $tpl$ (line 7) to compute its bounds while executing \textit{no ops} on other POMDPs that are not in $tpl$ (line 8). It then updates the bounds on the value of the full agent POMDP (line 9). The algorithm then removes the tuples for which the upper-bounds are less than the lower-bound of the agent POMDP and returns the action from the $tpl$ with the highest upper-bound (lines 10-11). 

The size of the sub-problems and their bounds gets updated as the truncated horizon $h$ increases. Function \texttt{RecomputeTuples} updates the $tpls$ set as the number of tasks that the robot can attend to within the horizon increases from $k$ to $k'=k+1$. For each $tpl \in tpls$, a member of $tpl_l$ is removed and added to its $tpl_c$ set. We consider removing any element from the $tpl_l$ set to generate all possible new tuples. This is to ensure that the optimality guarantees hold as we increase the truncated horizon $h$.

\begin{algorithm}
\caption{Multi-task POMDP with Adaptive Horizon}
\label{alg:acpomdp}
\SetKwFunction{FMain}{}
\SetKwProg{Fn}{InitializeTuples}{}{}
\Fn{\FMain{$P$,$h$,$H$}}{
        $k$,$k^*$  $\gets$ the maximum \# tasks the robot can attend to within $h$ and $H$;  $T$ $\gets$ $\{tpl: tpl \in \euscr{P} (P),|tpl| = k^*\}$ \\
        $tpls'$ $\gets$ $\{(tpl_c,tpl_l): tpl_u \in T, tpl_c \in \euscr{P} (tpl_u),|tpl_c| = k,tpl_l = tpl_u \setminus  tpl_c \}$ \\
	    \Return $tpls'$
}
\SetKwFunction{FMain}{}
\SetKwProg{Fn}{SelectAction}{}{}
\Fn{\FMain{$P$,$h$,$H$,$tpls$}}{
	\For{${tpl} \in tpls$}{
            ($\munderbar{V}_{tpl}$,$\bar{V}_{tpl}$) $\gets$ TruncatedAgentPOMDP($h$,$H$,$tpl$) \\
            ($\munderbar{U}_{tpl},\bar{U}_{tpl}$) $\gets$ $(\munderbar{V}_{tpl},\bar{V}_{tpl})+{\sum_{{q\in P \setminus  {{{tpl}_u}}}}} V_q^{n}$ \\
            $\bar{V}_P = \max(\bar{V}_P,\bar{U}_{tpl})$;    $\munderbar{V}_P = \max(\munderbar{V}_P,\munderbar{U}_{tpl})$
	}
    $tpls$ $\gets$ $\{tpl: tpl \in tpls, \bar{U}_{tpl} \geq \munderbar{V}_P \}$ \\
	$a_{best} \gets$ action from the $tpl$ with highest $\bar{U}_{tpl}$ \\
	\Return $a_{best}$,$tpls$,$\munderbar{V}_P$,$\bar{V}_P$
	}
\SetKwFunction{FMain}{}
\SetKwProg{Fn}{RecomputeTuples}{}{}
\Fn{\FMain{$h$,$tpls$}}{
    $k$,$k'$ $\gets$ the maximum \# tasks the robot can attend to within $h-1$ and $h$; $tpls'$ $\gets$ $tpls$ \\
    \If{$k \neq k'$}{
         $tpls'$ $\gets$ $\{(tpl_c \cup \{p\},tpl_l \setminus \{p\}): tpl \in tpls, p \in tpl_l \}$ \\
    }
	\Return $tpls'$
	}
\end{algorithm}

Other improvements that can be added to Alg.~\ref{alg:acpomdp} to further expedite planning include 1) for a given tuple $tpl$, if $\bar{V}_{tpl} = \munderbar{V}_{tpl}$, we do not need to recompute the $tpl$'s bounds as it is already optimal, 2) the tuples can be processed in the decreasing order of their upper-bounds so if the updated $\munderbar{V}_{P}$ is greater than the next tuple's upper-bound, the tuple and the remaining tuples in the list can be discarded, and 3) if desirable, a timeout condition\footnote{Our adaptive horizon algorithm can generate and improve a solution in an anytime fashion until the optimal solution is achieved.} can also be added to the conditions on line 4 of Alg.~\ref{alg:main} to ensure online performance.

\section{Optimality Proofs}

\noindent In this section, we first prove that agent-POMDP-AH computes an optimal solution. We then prove that multi-task-AH finds the same solution as  agent-POMDP-AH. We discuss both the proofs and the intuition behind them.
The proofs use the independent tasks definition, as stated in~\cite{mohseni2020efficient}. 
\\
\\
\noindent {\textbf{Notation required for understanding the intuition behind the proofs (mostly borrowed from~\cite{mohseni2020efficient}):}}
\begin{itemize}
\setlength\itemsep{0.07cm}
\item {$V^*_{p,t}$:} the optimal value of the client POMDP $p$ at time $t$. 

\item {$V^{n}_{p,t}$:} the value of following a trajectory of \textit{no ops} for the client POMDP $p$ at time $t$.

\item {$V^*_{P,t}$:} the optimal value of the agent POMDP created from the POMDPs in $P$ at time $t$ (Eq.~\ref{eq:agent_POMDP_value}).




\item {$V^*_{tpl,t}$:} the optimal value of the agent POMDP created from only the client POMDPs in $tpl$ at time $t$.

\item {$\munderbar{V}^{h}_{tpl,t}(b_{tpl})$,$\bar{V}^{h}_{tpl,t}(b_{tpl})$:} the lower and upper-bound on the value of a belief node $b_{tpl}$ in the belief tree of a truncated agent POMDP created only from the members of $tpl$ till $h$. The bounds on the values of the fringe nodes of the truncated belief tree are computed using Eq.~\ref{eq:LB} and Eq.~\ref{eq:UB}.

\end{itemize}

\hfill \break
\noindent {\textbf{More notation required for understanding the proofs (mostly borrowed from~\cite{mohseni2020efficient})}}
\begin{itemize}
\item {$\mathbb{B}^{*}$:} this refers to the Bellman operator.

\item {$A_{tpl}$:} only considers the actions associated with the POMDPs in $tpl$ and performs \textit{no op} on the other POMDPs (same as Eq.~\ref{eq:A_valid}, but the union is over $tpl$, not $P$). 

\item $Q^{*}_{p,t}(b,a)$: the optimal value of the client POMDP $p$ at time $t$ for belief $b$ and action $a$.

\item {$U^*_{tpl,t}$:} the optimal value of the agent POMDP built from $P$ with the action set $A_{tpl}$. Intuitively, $U^*_{tpl,t}$ considers both the value of the POMDPs in $tpl$ ($V^*_{tpl,t}$) and the value of executing \textit{no ops} on the ones that are not in $tpl$.
\end{itemize}

\subsection{Lower and upper-bound} 

\noindent We show that the bound computations are valid (Lem. $1$ and $2$) and monotone (Lem. $3$ and $4$). The monotonicity property assures that the lower and upper-bounds on the value of a belief node does not change or improves after each iteration of the algorithm (increase in the truncated horizon $h$). The bounds on the value of the fringe nodes are computed for the remaining horizon $H-h$ (or $\infty$ in the infinite-horizon case) using Eq.~\ref{eq:LB} and Eq.~\ref{eq:UB}. The bounds for the non-fringe nodes are computed by propagating the bound computations of the fringe nodes up to the root belief node. We do not make any assumptions regarding the maximum possible horizon in the bound computations, thus the lemmas also hold for the infinite-horizon problems with discounting. We use mathematical induction to prove Lem. 1 to 4.

\hfill \break
\noindent\textbf{Lemma 1}\hspace{0.2cm} \textit{Eq.~\ref{eq:LB} provides a lower-bound on the value of a tuple $tpl=({tpl}_c , {tpl}_l)$ where ${tpl}_u = {tpl}_c \cup {tpl}_l$.}

\small  \begin{equation} \label{eq:LB}
\begin{split}
\munderbar{V}_{tpl,t}(b_{tpl}) = \max_{p\in tpl_u} {\Bigl [}V^{*}_{p,t}(b_p) + \smashoperator{\sum\limits_{{q\in tpl_u \setminus \{p\}}}} V^{n}_{q,t}(b_q){\Bigl ]} \leq V^{*}_{tpl,t}(b_{tpl})
\end{split}\raisetag{0.8\baselineskip}
\end{equation}  \normalsize

\hfill \break
\noindent\textbf{Intuition behind proof:}\hspace{0.2cm} Let us consider that only one task from $tpl_u$, $p\in tpl_u$, can be executed till the full horizon ($V^{*}_{p}$), and we perform \textit{no ops} on the other tasks (${\sum} V^{n}_{q}$). The best task will then be selected as the lower-bound on $V^{*}_{tpl}$, $\max\limits_{p} {[}V^{*}_{p} + \sum V^{n}_{q}{]}$. 

\hfill \break
\noindent\textbf{Proof:}\hspace{0.2cm} The proof goes by mathematical induction. For $h'=1$, if $\forall p \in P,  V^*_{p,0}(b_p) = 0$, Eq.~\ref{app_eq:h1} follows from Eq.~\ref{eq:agent_POMDP_value}:


\small  \begin{equation}
\label{app_eq:h1}
\begin{split}
V^*_{tpl,1}(b_{tpl}) &=\overbrace{\smashoperator{\max_{p\in {tpl_u}}}{\Bigl [}\smashoperator{\max_{a\in {A_{p}}}}}^{{\max_{a\in {A_{tpl_u}}}}}{\Bigl [}\sum\limits_{{i\in tpl_u}}\sum\limits_{{s\in S_i}}b_i(s)R_i(s,a[i]){\Bigr ]}{\Bigr ]} \\ 
& = \smashoperator{\max_{p\in {tpl_u}}}{\Bigl [}V^*_{p,{1}}(b_{p}) + \smashoperator{\sum\limits_{{q\in tpl_u \setminus \{p\}}}} V^{n}_{q,1}(b_q){\Bigr ]}
\end{split}\raisetag{2\baselineskip}
\end{equation}  \normalsize

\noindent If $h'=t-1$, we assume Eq.~\ref{app_eq:parallel_LB_max} and consequently Eq.~\ref{app_eq:parallel_LB} and show that they both hold for $h'=t$. 


\small  \begin{equation}
\label{app_eq:parallel_LB_max}
\begin{split}
V^*_{tpl,t-1}(b_{tpl}) & \geq \smashoperator{\max_{p\in {tpl_u}}}{\Bigl [}V^*_{p,{t-1}}(b_{p}) + \smashoperator{\sum\limits_{{q\in tpl_u \setminus \{p\}}}} V^{n}_{q,t-1}(b_q){\Bigr ]}
\end{split}
\end{equation}  \normalsize

\small  \begin{equation} \label{app_eq:parallel_LB}
\begin{split}
\forall {p \in tpl_u}: V^*_{tpl,t-1}(b_{tpl}) & \geq V^*_{p,{t-1}}(b_{p}) + \smashoperator{\sum\limits_{{q\in tpl_u \setminus \{p\}}}} V^{n}_{q,t-1}(b_q)
\end{split} \raisetag{2\baselineskip}
\end{equation}  \normalsize

\noindent We  expand Eq.~\ref{eq:agent_POMDP_value} as follows ($b_{tpl}$ or $b$):
\small  \begin{equation} \label{app_eq:LB_tpl}
\begin{split}
&V^*_{tpl,t}(b) =\max_{a\in {A_{tpl_u}}}{\Bigl [}\sum\limits_{{i\in tpl_u}}\sum\limits_{{s\in S_i}}b_i(s)R_i(s,a[i]) \\
&+ \gamma {\sum\limits_{z_q\in Z_q}\Pr(z_q|b_q,a_q)\ldots \smashoperator{\sum\limits_{z_r\in Z_r}}\Pr(z_r|b_r,a_r)}V^*_{tpl,t-1}(b_z^a){\Bigr ]} \\
\end{split} \raisetag{3\baselineskip}
\end{equation}  \normalsize

\noindent We substitute Eq.~\ref{app_eq:parallel_LB} in Eq.~\ref{app_eq:LB_tpl}. Given the independence assumption, for a specific $Z_i$, we can marginalize out the sum over $Z_j$s ($j \neq i$). $\forall {p \in tpl_u}$, we obtain:

\small  \begin{equation} \label{app_eq:parallel_d}
\begin{split}
 & V^{*}_{tpl,t}(b)  \geq \max_{a\in {A_{tpl}}}{\Bigl [}Q^{*}_{p,t-1}(b_p,a[p]) + \overbrace{\smashoperator{\sum\limits_{{q\in tpl_u \setminus \{p\}}}}Q^{n}_{q,t-1}(b_q,a[q])}^{Q_{noop}}{\Bigr ]} \\
& \geq \max_{a\in {A_{p}}}{\Bigl [}Q^{*}_{p,t-1}(b_p,a[p]) + {Q_{noop}}{\Bigr ]}  \geq V^*_{p,{t}}(b_{p}) + \smashoperator{\sum\limits_{{q\in tpl_u \setminus \{p\}}}} V^{n}_{q,t}(b_q)
\end{split}\raisetag{0.7\baselineskip}
\end{equation}  \normalsize

\noindent Thus, Eq.~\ref{app_eq:parallel_full} holds for every $h'=t$.

\small  \begin{equation}
\label{app_eq:parallel_full}
\begin{split}
V^*_{tpl,t}(b) & \geq \smashoperator{\max_{p\in {tpl_u}}}{\Bigl [}V^*_{p,{t}}(b_{p}) + \smashoperator{\sum\limits_{{q\in tpl_u \setminus \{p\}}}} V^{n}_{q,t}(b_q){\Bigr ]}
\end{split}\raisetag{4\baselineskip}
\end{equation}  \normalsize

\hfill \break
\noindent\textbf{Lemma 2}\hspace{0.2cm} \textit{Eq.~\ref{eq:UB} provides an upper-bound on the value of a tuple $tpl=({tpl}_c , {tpl}_l)$.}

\small  \begin{equation} \label{eq:UB}
\begin{split}
\bar{V}_{tpl,t}(b_{tpl}) = \smashoperator{\sum\limits_{{p\in tpl_u}}} V^{*}_{p,t}(b_p) \geq V^{*}_{tpl,t}(b_{tpl})
\end{split}\raisetag{0.2\baselineskip}
\end{equation}  \normalsize

\noindent\textbf{Intuition behind proof:}\hspace{0.2cm} The idea behind the upper-bound computation is to assume that the robot can attend to all the tasks in $tpl$, $p \in tpl_u$, in parallel ($\sum V^{*}_{p}$). We only have one robot, so this is an upper-bound on $V^{*}_{tpl}$.

\hfill \break
\noindent\textbf{Proof:}\hspace{0.2cm} Similar to Lem. 1, the proof goes by mathematical induction. For $h'=1$, the following equation holds.

\vspace{-0.3cm}
\small  \begin{equation}
\begin{split}
& V^*_{tpl,1}(b_{tpl}) =\smashoperator{\max_{a\in {A_{tpl_u}}}}{\Bigl [}\sum\limits_{{i\in tpl_u}}\sum\limits_{{s\in S_i}}b_i(s)R_i(s,a[i]){\Bigr ]} \\ 
& \leq \sum\limits_{{i\in tpl_u}}\smashoperator{\max_{a\in {A_{tpl_u}}}}{\Bigl [}\sum\limits_{{s\in S_i}}b_i(s)R_i(s,a[i]){\Bigr ]}  = 
\smashoperator{\sum\limits_{{i\in tpl_u}}} V^{*}_{i,1}(b_i)
\end{split}\raisetag{2\baselineskip}
\end{equation}  \normalsize
\vspace{-0.1cm}


\noindent We assume  Eq.~\ref{app_eq:parallel_UB} holds for $h'=t-1$ ($p,q,r,\ldots \in tpl_u$) and show that it also holds for $h'=t$.


\small  \begin{equation} \label{app_eq:parallel_UB}
\begin{split}
V^{*}_{tpl,t-1}(b) \leq V^{*}_{p,{t-1}}(b_{p}) + \ldots + V^{*}_{q,{t-1}}(b_{q}) + \ldots + V^{*}_{r,{t-1}}(b_{r})
\end{split}\raisetag{-0.15\baselineskip}
\end{equation}  \normalsize

\noindent Similar to Lem. 1, Eq.~\ref{app_eq:parallel_UB} is substituted in Eq.~\ref{app_eq:LB_tpl}, and simplified to obtain Eq.~\ref{app_eq:parallel_UB_t}. Thus, Eq.~\ref{eq:UB} holds for every $h'=t$.

\small  \begin{equation} \label{app_eq:parallel_UB_t}
\begin{split}
V^{*}_{tpl,t}(b) & \leq \max_{a\in {A_{tpl_u}}}{\Bigl [}\sum\limits_{{i\in tpl_u}}Q^{*}_{i,t}(b_i,a[i]){\Bigr ]} \\
& \leq \sum\limits_{{i\in tpl_u}} \max_{a\in {A_{tpl_u}}}Q^{*}_{i,t}(b_i,a[i]) = \smashoperator{\sum\limits_{{p\in tpl_u}}} V^*_{p,t}(b_p)
\end{split}\raisetag{8\baselineskip}
\end{equation}  \normalsize

\hfill \break
\noindent\textbf{Lemma 3}\hspace{0.2cm} \textit{The lower-bound computation is monotone.}



\small  \begin{equation} \label{eq:LB_rel}
\begin{split}
\munderbar{V}^{h}_{tpl,t}(b_{tpl}) \leq \munderbar{V}^{h'}_{tpl,t}(b_{tpl}) \\
\text{where } h < h' \text{ and } h,h'\leq H
\end{split}\raisetag{0.2\baselineskip}
\end{equation}  \normalsize


\noindent In both $\munderbar{V}^{h}_{tpl,t}$ and $\munderbar{V}^{h'}_{tpl,t}$'s computations, the belief tree is built till horizon $h$. To compute $\munderbar{V}^{h}_{tpl,t}$, the lower-bound on the value of the fringe belief nodes at horizon $h$ are computed using Eq.~\ref{eq:LB} and are propagated up the belief tree. To compute $\munderbar{V}^{h'}_{tpl,t}$, the algorithm expands the  tree for $d$ more steps, $h'=h+d$, and then uses Eq.~\ref{eq:LB} to compute the lower-bound for the fringe nodes at depth $h+d$ and propagates the bounds up the belief tree. In both cases the lower-bound on the value of the fringe nodes are computed using Eq.~\ref{eq:LB} till the full horizon $H$.
This  property guarantees that as the truncated horizon increases, from $h$ to $h'$ ($h < h'$), the lower-bound on the value of a certain fringe node at horizon $h$ and consequently the non-fringe nodes are non-decreasing. 

\hfill \break
\noindent\textbf{Intuition behind proof:}\hspace{0.2cm} The main difference between $\munderbar{V}^{h}(b')$ and $\munderbar{V}^{h'}(b')$ for a certain belief node $b'$ at depth $h$ (or horizon $h$) is that the former uses the trivial lower-bound estimate for the node, but the latter does more computation to expand the belief tree further before using a similar trivial lower-bound estimate for the nodes at depth $h+d$. To compute the lower-bound for a fringe node at depth $h$, $\munderbar{V}^{h}(b')$, the algorithm assumes that from there on till $H$, only one task can be executed and \textit{no ops} are executed on the other tasks (one possible solution). So, expanding the belief tree (exhaustive search) for $d$ more steps till horizon $h+d$ to compute $\munderbar{V}^{h'}(b')$ will only find the same or a better solution than achieving a single task. \emph{I.e.}, the lower-bound on $b'$ is non-decreasing as we increase the horizon.

\hfill \break
\noindent\textbf{Proof:}\hspace{0.2cm} For a certain leaf node $b$ at horizon $h$, we compare its lower-bound  when the truncated agent POMDP is built till $h$ against when it is built till $h'$. The proof goes by mathematical induction. First, we show that 
${\munderbar{V}_{tpl,H-h} \leq \mathbb{B}^{*}} \munderbar{V}_{tpl,H-h-1}$ holds for $d=1$. We proved this previously when we substitute Eq.~\ref{app_eq:parallel_LB} in Eq.~\ref{app_eq:LB_tpl} to get Eq.~\ref{app_eq:parallel_d}, thus:


\small  \begin{equation} \label{app_eq:parallel_a}
\begin{split}
& V^{*}_{tpl,H-h} = {\mathbb{B}^{*}}  V^{*}_{tpl,H-h-1} 
\geq {\mathbb{B}^{*}}  \munderbar{V}_{tpl,H-h-1} \\
&\geq \smashoperator{\max_{p\in {tpl_u}}}{\Bigl [}V^*_{p,{H-h}} + \smashoperator{\sum\limits_{{q\in tpl_u \setminus \{p\}}}} V^{n}_{q,H-h}{\Bigr ]} = \munderbar{V}_{tpl,H-h}
\end{split}\raisetag{8\baselineskip}
\end{equation}  \normalsize

\noindent Now, we assume that for $h'=h+d$, the following holds for the belief node $b$: ${\munderbar{V}_{tpl,H-h}\leq \mathbb{B}^{*}_{d}} \munderbar{V}_{tpl,H-h-d}$, and we prove that the same equation also holds if $h'=h+d+1$. 
For a certain belief $b$, both ${\munderbar{V}_{tpl,H-h}\leq \mathbb{B}^{*}} \munderbar{V}_{tpl,H-h-1}$ and ${\munderbar{V}_{tpl,H-h} \leq \mathbb{B}^{*}_{d}} \munderbar{V}_{tpl,H-h-d}$ hold, thus the following equation holds for $h'=h+d+1$:

\small  \begin{equation} \label{app_eq:parallel_b}
\begin{split}
V^{*}_{tpl,H-h} & \geq {\mathbb{B}^{*}_{d}}{\Bigl [}{\mathbb{B}^{*}} \munderbar{V}_{tpl,H-h-d-1}{\Bigl ]} \geq {\mathbb{B}^{*}_{d}} \munderbar{V}_{tpl,H-h-d} \\ 
& \geq \smashoperator{\max_{p\in {tpl_u}}}{\Bigl [}V^*_{p,{H-h}} + \smashoperator{\sum\limits_{{q\in tpl_u \setminus \{p\}}}} V^{n}_{q,H-h}{\Bigr ]= {\munderbar{V}_{tpl,H-h}}}
\end{split}\raisetag{1\baselineskip}
\end{equation}  \normalsize



\hfill \break
\noindent\textbf{Lemma 4}\hspace{0.2cm} \textit{The upper-bound computation is monotone.}

\small  \begin{equation} \label{eq:UB_rel}
\begin{split}
\bar{V}^{h}_{tpl,t}(b_{tpl}) \geq \bar{V}^{h'}_{tpl,t}(b_{tpl}) \\
\text{where } h < h' \text{ and } h,h'\leq H
\end{split}\raisetag{0.2\baselineskip}
\end{equation}  \normalsize
\vspace{-0.35cm}

\noindent This  property guarantees that 
as the horizon increases, from $h$ to $h'$, the upper-bound on the value of a certain fringe node and consequently the non-fringe nodes are non-increasing.

\hfill \break
\noindent\textbf{Intuition behind proof:}\hspace{0.2cm} Similar to the intuition we gave for the lower-bound's monotonicity, for a certain belief node $b'$ at depth $h$, $\bar{V}^{h}(b')$ estimates the upper-bound  by assuming that all the tasks can be performed in parallel. However, $\bar{V}^{h'}(b')$ expands the belief tree for $d$ more steps before assuming that all the tasks can be performed in parallel. Thus, given that $\bar{V}^{h'}(b')$ uses the Bellman equation during the $d$ steps, it has a better estimate of the upper-bound than the assumption that all the tasks can be attended to in parallel during that $d$ steps as assumed in $\bar{V}^{h}(b')$. \emph{I.e.}, as the horizon increases and more of the belief tree is expanded, the upper-bound on the value of $b'$ improves (\emph{i.e.}, is non-increasing).

\hfill \break
\noindent\textbf{Proof:}\hspace{0.2cm} Similar to Lem. 3's proof, the proof goes by mathematical induction. First, we show that 
${\mathbb{B}^{*}} \bar{V}_{tpl,H-h-1} \leq \bar{V}_{tpl,H-h}$ holds for $d=1$. We proved this previously when we substitute Eq.~\ref{app_eq:parallel_UB} in Eq.~\ref{app_eq:LB_tpl} to get Eq.~\ref{app_eq:parallel_UB_t}, thus:

\small  \begin{equation} \label{app_eq:parallel_c}
\begin{split}
& V^{*}_{tpl,H-h} = {\mathbb{B}^{*}}  V^{*}_{tpl,H-h-1}  \leq \max_{a\in {A_{tpl_u}}}{\Bigl [}\sum\limits_{{i\in tpl_u}}Q^{*}_{i,H-h-1}{\Bigr ]} \\
& \leq \sum\limits_{{i\in tpl_u}} \max_{a\in {A_{tpl_u}}}Q^{*}_{i,H-h-1} = \smashoperator{\sum\limits_{{i\in tpl_u}}} V^*_{i,H-h} = \bar{V}_{tpl,H-h}
\end{split}\raisetag{1\baselineskip}
\end{equation}  \normalsize

\noindent We assume that for the belief node $b$ and $h'=h+d$,  ${\mathbb{B}^{*}_{d}} \bar{V}_{tpl,H-h-d} \leq \bar{V}_{tpl,H-h}$ holds, and we prove it also holds if $h'=h+d+1$. 
We know both ${\mathbb{B}^{*}} \bar{V}_{tpl,H-h-1} \leq \bar{V}_{tpl,H-h}$ and ${\mathbb{B}^{*}_{d}} \bar{V}_{tpl,H-h-d} \leq \bar{V}_{tpl,H-h}$ hold, thus,

\small  \begin{equation} \label{app_eq:parallel_5}
\begin{split}
V^{*}_{tpl,H-h} \leq {\mathbb{B}^{*}_{d}}{\mathbb{B}^{*}} \bar{V}_{tpl,H-h-d-1} \leq {\mathbb{B}^{*}_{d}} \bar{V}_{tpl,H-h-d} \leq
\bar{V}_{tpl,H-h}
\end{split}\raisetag{-0.1\baselineskip}
\end{equation}  \normalsize


\noindent In summary, we proved that the bound computations are valid and monotone; thus if $tpl = (P,\emptyset)$, Lem. 1 to 4 prove the optimality of agent-POMDP-AH. Given the iterative nature of the horizon, in the worst case, the agent-POMDP-AH approach reaches the full horizon $H$ and obtains the same solution as the agent-POMDP-FH approach.

\subsection{Multi-task-AH} 

\noindent We prove Alg.~\ref{alg:acpomdp} is optimal. We assume $k^*$ and $k$ are the maximum number of tasks that the robot can attend to within $H$ and the truncated horizon $h$ respectively. $\hat{{V}}_{P}$ denotes the value of the agent POMDP under such assumptions, referred to as \textit{limited tasks assumption}. 



\hfill \break
\noindent\textbf{Lemma 5}\hspace{0.2cm} \textit{The lower and upper-bounds on the value of the agent POMDP created from the set $P$, $\hat{\munderbar{V}}_{P}$ and $\hat{\bar{V}}_{P}$, can be computed by Eq.~\ref{eq:k_tuple_LB} and Eq.~\ref{eq:k_tuple_UB}  respectively where $tpls = \{tpl \in \euscr{P} (P):|tpl| = k^*\}$, and the bounds are monotone. (proof of \texttt{SelectAction} function in Alg.~\ref{alg:acpomdp})} 

\small  \begin{equation} \label{eq:k_tuple_LB}
\begin{split}
&\hat{\munderbar{V}}_{P,t} (b) = \max_{tpl \in tpls} (\munderbar{V}_{tpl,t}(b_{tpl})+ \smashoperator{\sum\limits_{{q \in P \setminus tpl_u}}} V^{n}_{q,t}(b_q)) \leq \hat{{V}}^*_{P,t} (b) \\
\end{split}\raisetag{\baselineskip}
\end{equation}  \normalsize

\small  \begin{equation} \label{eq:k_tuple_UB}
\begin{split}
&\hat{\bar{V}}_{P,t} (b) = \max_{tpl \in tpls} (\bar{V}_{tpl,t}(b_{tpl})+ \smashoperator{\sum\limits_{{q \in P \setminus tpl_u}}} V^{n}_{q,t}(b_q)) \geq \hat{{V}}^*_{P,t} (b) \\
\end{split}
\end{equation}  \normalsize

\noindent\textbf{Intuition behind proof:}\hspace{0.2cm} In~\cite{mohseni2020efficient}, we proved that finding the optimal values of all $tpl \in tpls$ ($V^{*}_{tpl}$) while performing \textit{no ops} on the other POMDPs ($\sum{V^{n}_{q}}$)  and selecting the best V-value, $\max_{tpl \in tpls}{(V^{*}_{tpl}+\sum{V^{n}_{q}})}$, provides the optimal solution to the agent POMDP. We proved in Lem. 1 to 4 that the lower and upper-bounds on $V^{*}_{tpl}$ are valid and monotone. The validity and monotonicity of $\hat{\munderbar{V}}_{P}$ and $\hat{\bar{V}}_{P}$ then simply follow from the validity and monotonicity of $\munderbar{V}_{tpl}$ and $\bar{V}_{tpl}$.

\hfill \break
\noindent\textbf{Proof:}\hspace{0.2cm} We show that the bounds are valid and then argue why they are also monotone. From~\cite{mohseni2020efficient}, we know:
\vspace{-0.1cm}
\small  \begin{equation} \label{app_eq:k_tuple_optimal}
\begin{split}
\hat{V}^*_{P,t} (b) & = \max_{tpl \in tpls} U^{*}_{{tpl,t}}(b) 
\end{split}
\end{equation}  \normalsize

\vspace{-0.1cm}
\small  \begin{equation} \label{app_eq:k_tuple_value}
\begin{split}
&U^{*}_{{tpl,t}}(b) = V^*_{{tpl,t}}(b_{{tpl}}) + \smashoperator{\sum\limits_{{q\in P \setminus {tpl}}}} V^{n}_{q,t}(b_q)
\end{split}
\end{equation}  \normalsize
\noindent We proved in Lem. $1$ and $2$ that $\munderbar{V}_{tpl,t}(b_{tpl})  \leq V^{*}_{tpl,t}(b_{tpl})$ and $\bar{V}_{tpl,t}(b_{tpl})  \geq V^{*}_{tpl,t}(b_{tpl})$ respectively. Thus, $\munderbar{U}_{{tpl,t}}(b)$ and $\bar{U}_{{tpl,t}}(b)$ computed by substituting  ${V}^{*}_{tpl,t}(b_{tpl})$ by $\munderbar{V}_{tpl,t}(b_{tpl})$ and $\bar{V}_{tpl,t}(b_{tpl})$ in Eq.~\ref{app_eq:k_tuple_value} are lower and upper-bounds on ${U^*}_{{tpl,t}}(b)$. We substitute $\munderbar{U}_{{tpl,t}}(b)$ and $\bar{U}_{{tpl,t}}(b)$ in Eq.~\ref{app_eq:k_tuple_optimal} to prove Eq.~\ref{eq:k_tuple_LB} and Eq.~\ref{eq:k_tuple_UB}. Given that the bound computations for $V^{*}_{tpl,t}$ are monotone, $\smashoperator{\sum} V^{n}_{q,t}(b_q)$ does not change for a given tuple as we increases the horizon, and the $max$ operator does not change the monotonicity of $\munderbar{U}_{{tpl,t}}$ and $\bar{U}_{{tpl,t}}$, $\hat{\munderbar{V}}_{P,t}$ and $\hat{\bar{V}}_{P,t}$  are monotone.

\hfill \break
\noindent\textbf{Lemma 6}\hspace{0.2cm}\textit{  Alg.~\ref{alg:acpomdp}  converges to the optimal solution of the agent POMDP in both finite horizon problems without discounting and infinite horizon problems with discounting.}

\hfill \break
\noindent\textbf{Intuition behind proof:}\hspace{0.2cm} In Lem. 1 to 5, we proved that dividing the agent POMDP into subtasks ($tpl \in tpls$) and computing the lower and upper-bounds for all the tuples in $tpls$  provide valid and monotone bounds on the value of the agent POMDP. In those lemmas, we assumed that  $k = k^*$, \emph{i.e.}, a combined model of all the $k^*$ tasks is expanded till the truncated horizon $h$ even though we know that the robot can only attend to $k$ tasks within $h$. Differently, in Alg.~\ref{alg:acpomdp}, to efficiently solve each $tpl$ for a small truncated horizon $h$, we only consider subsets of size $k$, but compute the bounds on all the $k^*$ POMDPs in $tpl$, so $k<k^*$. Both cases $k=k^*$ and $k < k^*$ use the same lower and upper-bound computations and have the same $h$ as their truncated horizon. However, in the former we perform the tree expansion on a combined model built from all the POMDPs in the $tpl$ set, but in the latter we consider all combinations of the tasks with size $k$ out of the POMDPs in $tpl$ and perform the tree expansion on those only. The proof uses the same idea as~\cite{mohseni2020efficient}. It uses the assumption that within a certain horizon $h$, only $k$ tasks can be attended to, so if we consider all combinations of $k$ tasks out of the members of the $tpl$ ($tpl_u$), we will get the same solution as the combined model of all the tasks in $tpl$.
Given that the bound computations are the same in both cases, when $k < k^*$, we get the same solution as when $k=k^*$ (proof for line 3 in Alg.~\ref{alg:acpomdp} and the \texttt{RecomputeTuples} function), and Alg.~\ref{alg:acpomdp} computes  valid and monotone bounds on the value of the agent POMDP.

  
\hfill \break
\noindent\textbf{Proof:}\hspace{0.2cm} 
In Lem. 1 to 5, we proved that dividing the agent POMDP into subtasks ($tpl \in tpls$) and computing the lower and upper-bounds for all the members of $tpls$  provide valid and monotone bounds on the value of the agent POMDP. In these lemmas, we assumed that $k = k^*$, so line 3 of Alg.~\ref{alg:acpomdp} would become $tpls' = \{(tpl_c,tpl_l): tpl \in tpls, tpl_c =tpl, tpl_l=\emptyset \}$, and the \texttt{RecomputeTuples} function would not change the $tpls$ set. However, the benefits of our approach are manifested when the truncated horizon $h$ is smaller than the full planning horizon $H$, and consequently  $k < k^*$. In Alg.~\ref{alg:acpomdp}, we divide each $tpl$ into two sets, $tpl_c$ with $k$ tasks and $tpl_l$ with $k^*-k$ tasks  (all possible combinations of $k$ tasks out of $k^*$ tasks), perform the tree expansion for the POMDPs in $tpl_c$ while executing \textit{no ops} on the members of $tpl_l$, and compute the bounds on all members of $tpl_u = tpl_c \cup tpl_l$. When $k < k^*$, if we prove that by using this approach, we get the same solution as when $k=k^*$, we prove that Alg.~\ref{alg:acpomdp} computes valid and monotone bounds on the value of the agent POMDP.



Notice that the only difference between $k=k^*$ and $k < k^*$ is that in the former we perform the tree expansion on all POMDPs in the $tpl$ set, but in the latter we consider all combinations of tasks with size $k$ for the $tpl_c$ set and perform the tree expansion on the POMDPs in $tpl_c$ only. The lower and upper-bound computations are the same in both cases.

We use the same idea as~\cite{mohseni2020efficient}, Lem. 2 and Asm. 1 in  \cite{mohseni2020efficient}. For a set of tasks called $tpl$ and the maximum number of tasks that the robot can attend to within the horizon $h$ ($k$), the robot can optimally solve the combined model of all tasks by considering all subsets of tasks of size $k$ ($tpl_c$ with size $k$). 
Given the independence between the tasks and the limited tasks assumption, Eq.~\ref{app_eq:div} was proved in~\cite{mohseni2020efficient} for $tpl=P$ and $h=H$ ($k^*$ tasks). Same deductions also apply here to prove Eq.~\ref{app_eq:div}. 

\small  \begin{equation} \label{app_eq:div}
\begin{split}
&\hat{V}^{*}_{tpl,t} (b) = \max_{tpl' \in tpls'} {V}^{*}_{{tpl'_c,t}}(b_{{tpl'_c}}) + \smashoperator{\sum\limits_{{q\in {tpl'_l}}}} V^{n}_{q,t}(b_q)
\end{split}
\end{equation}  \normalsize
\vspace{-0.35cm}
\scriptsize  \begin{equation*}
\begin{split}
tpls' = \{(tpl_c,tpl_l):  tpl_c \in \euscr{P} (tpl),|tpl_c| = k,tpl_l = tpl_u \setminus  tpl_c \} 
\end{split}
\end{equation*}  \normalsize

This explains why dividing $tpl$ further into subsets of size $k$ (line 3 in Alg.~\ref{alg:acpomdp}) does not change the validity and monotonicity of the bounds and gives us the same bounds as if we were to build a combined model of  all the POMDPs in $tpl_u$. 

As we increase $h$, $k$ should also increase to ensure that Eq.~\ref{app_eq:div} is still valid. More specifically, we have to update each tuple in the $tpls'$ set ($tpl' \in tpls'$) to have $k+1$ POMDPs in $tpl_c$ and $k^*-k-1$ POMDPs in $tpl_l$. This is done by the \texttt{RecomputeTuples} function. The algorithm simply removes a POMDP from $tpl_l$ and adds it to the POMDPs in $tpl_c$ to create a new $tpl'_c$ set of size $k+1$ and a new $tpl'_l$ set of size $k^*-k-1$, $tpl' = (tpl'_c,tpl'_l)$. The algorithm considers removing any POMDP from $tpl_l$, to create all possible new tuples. Since the new $tpls'$ set satisfies the limited tasks assumption as we increase the horizon, Eq.~\ref{app_eq:div} holds.

Therefore, all parts of the algorithm preserve the optimality guarantees, and Alg.~\ref{alg:acpomdp}  computes  an  optimal  solution  for  the agent POMDP. In the worst case, the multi-task-AH approach reaches the full horizon $H$ and obtains the same solution as the multi-task-FH approach. In infinite-horizon problems with discounting, when $h \rightarrow \infty$, Alg.~\ref{alg:acpomdp} converges to the optimal solution of the agent POMDP with $H = \infty$.



\bibliographystyle{IEEEtran}

\bibliography{main}

\end{document}